\documentclass[nohyperref]{article}

\usepackage{microtype}
\usepackage{graphicx}
\usepackage{subfigure}
\usepackage{booktabs} % for professional tables
\usepackage{multirow}
\usepackage{amsmath,bm}

\usepackage{hyperref}

\usepackage[preprint]{icml2022}

\usepackage{amsmath}
\usepackage{amssymb}
\usepackage{mathtools}
\usepackage{amsthm}

\usepackage[capitalize,noabbrev]{cleveref}

\theoremstyle{plain}

\theoremstyle{definition}

\theoremstyle{remark}

\usepackage[textsize=tiny]{todonotes}

\begin{document}

\twocolumn[
% \icmltitle{Prompt-Guided Injection of Conformation to \\Pre-trained Protein Model}
\icmltitle{Prompt-Guided Injection of Conformation to Pre-trained Protein Model}

% It is OKAY to include author information, even for blind
% submissions: the style file will automatically remove it for you
% unless you've provided the [accepted] option to the icml2022
% package.

% List of affiliations: The first argument should be a (short)
% identifier you will use later to specify author affiliations
% Academic affiliations should list Department, University, City, Region, Country
% Industry affiliations should list Company, City, Region, Country

% You can specify symbols, otherwise they are numbered in order.
% Ideally, you should not use this facility. Affiliations will be numbered
% in order of appearance and this is the preferred way.
\icmlsetsymbol{equal}{*}

\begin{icmlauthorlist}
\icmlauthor{Qiang Zhang}{equal,zju,kczx,azft}
\icmlauthor{Zeyuan Wang}{equal,zju,kczx,azft}
\icmlauthor{Yuqiang Han}{zju,kczx,azft}
\icmlauthor{Haoran Yu}{zju,kczx}
\icmlauthor{Xurui Jin}{mr}
\icmlauthor{Huajun Chen}{zju,kczx,azft}
\end{icmlauthorlist}

\icmlaffiliation{zju}{College of Computer Science and Technology, Zhejiang University}
\icmlaffiliation{kczx}{Hangzhou Innovation Center, Zhejiang University}
\icmlaffiliation{azft}{AZFT Knowledge Engine Lab}
\icmlaffiliation{mr}{MindRank AI Ltd., China}
%\icmlcorrespondingauthor{Qiang Zhang}{qiang.zhang.cs@zju.edu.cn}
%\icmlcorrespondingauthor{Zeyuan Wang}{yuanzew@zju.edu.cn}
% \icmlcorrespondingauthor{Yuqiang Han}{hyq2015@zju.edu.cn}
% \icmlcorrespondingauthor{Haoran Yu}{yuhaoran@zju.edu.cn}
% \icmlcorrespondingauthor{Xurui Jin}{xurui@mindrank.ai}
\icmlcorrespondingauthor{Huajun Chen}{huajunsir@zju.edu.cn}

% You may provide any keywords that you
% find helpful for describing your paper; these are used to populate
% the "keywords" metadata in the PDF but will not be shown in the document
\icmlkeywords{Machine Learning, ICML}

\vskip 0.3in
]

% this must go after the closing bracket ] following \twocolumn[ ...

% This command actually creates the footnote in the first column
% listing the affiliations and the copyright notice.
% The command takes one argument, which is text to display at the start of the footnote.
% The \icmlEqualContribution command is standard text for equal contribution.
% Remove it (just {}) if you do not need this facility.

%\printAffiliationsAndNotice{}  % leave blank if no need to mention equal contribution
\printAffiliationsAndNotice{\icmlEqualContribution} % otherwise use the standard text.

\begin{abstract}
Pre-trained protein models (PTPMs) represent a protein with one fixed embedding and thus are not capable for diverse tasks.
For example, protein structures can shift, namely protein folding, between several conformations in various biological processes.
To enable PTPMs to produce informative representations, we propose to learn interpretable, pluggable, and extensible protein prompts as a way of injecting task-related knowledge into PTPMs.
In this regard, prior PTPM optimization with the {masked language modeling} task can be interpreted as learning a \emph{sequence prompt} (Seq prompt) that enables PTPMs to capture the {sequential dependency} between amino acids. 
To incorporate conformational knowledge to PTPMs, we propose an \emph{interaction-conformation prompt} (IC prompt) that is 
learned through back-propagation with the protein-protein interaction task.
As an instantiation, we present a conformation-aware pre-trained protein model that learns both the sequence and interaction-conformation prompts in a multi-task setting.
We conduct comprehensive experiments on nine protein datasets. Results show that using the Seq prompt does not hurt PTPMs' performance on sequence-related tasks while incorporating the IC prompt significantly improves PTPMs' performance on tasks where interaction conformational knowledge counts.
{Furthermore, the learned prompts can be combined and extended to deal with new protein tasks}.  
%This paper provides fresh insights into how to utilize prompts to enhance PTPMs for various downstream tasks.
\end{abstract}

\section{Introduction}

Proteins play an essential role in biological activities. As proteins are composed of sequences of amino acids, the chemical properties of amino acids cause complex dynamic 3D structures and determine the protein functions as a whole~\cite{Epstein1963TheGC}. 
One popular approach to deal with sequence data is pre-trained language models (PTLMs),
which have achieved excellent performance in language understanding~\cite{devlin-etal-2019-bert} and translation~\cite{2020t5} and dialogue systems~\cite{DBLP:journals/corr/abs-1907-05339}.
Inspired by that, researchers have developed pre-trained protein models (PTPMs), such as TAPE Transformer~\cite{tape2019}, ProtBERT~\cite{9477085}, and ESM-1b~\cite{rao2021transformer}, to predict protein structures and functions. PTPMs have achieved promising performance on various downstream tasks, such as secondary structure prediction~\cite{pdb}, affinity prediction~\cite{10.1093/nar/gkt1043}, and contact prediction~\cite{doi:10.1002/prot.25415}.  

\begin{figure}[t]
%\vskip 0.2in
\begin{center}
\centerline{\includegraphics[width=\columnwidth]{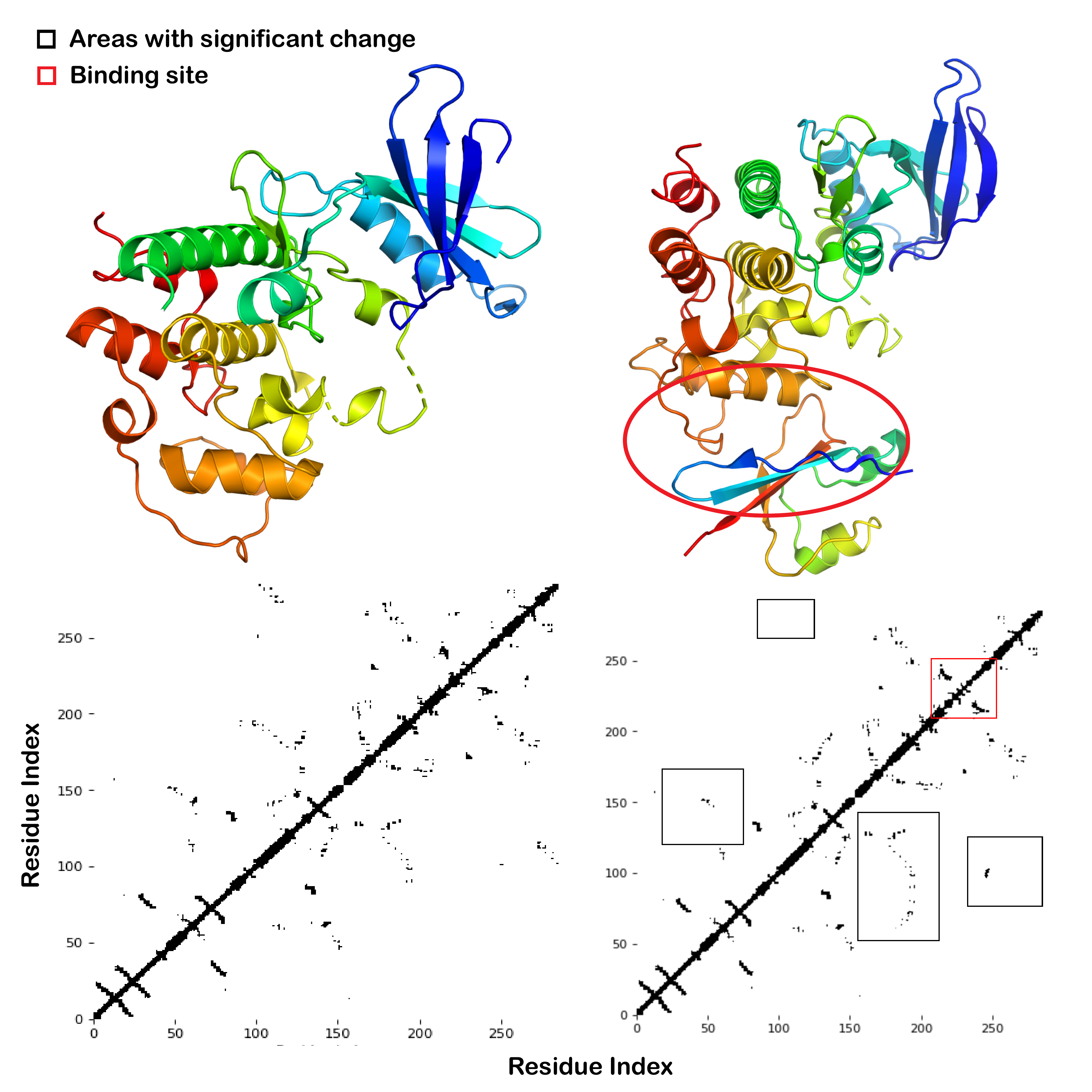}}
\vskip -0.1in
\caption{
The structure and contact map of protein CDK1 (pdbid: 4yc6). 
CDK1 is the only essential cell cycle CDK in human cells and is required for successful completion of M-phase. 
Here, we compare the native conformation and the interaction conformation with CSK1 protein (pdbid:2RSY). From the contact map, we can observe that there is a great difference between the two different conformations.}
\label{conformation}
\end{center}
\vskip -0.25in
\end{figure}

However, proteins are complex biological structures and have unique characteristics. 
One important difference between sentences and amino acid sequences is that sentences have static structures and semantics while proteins composed of amino acids are dynamic and can be observed with various 3D structures, which are called conformations~\cite{BU2011163}.
Figure~\ref{conformation} shows an example of different conformations and contact maps of the same protein. It has been reported that protein conformations are very sensitive and dynamic, significantly influenced by external factors and their specific function~\cite{Harper's-Illustrated-Biochemistry}. Therefore, it is inappropriate for existing PTPMs to use a single fixed embedding to represent a protein. 

This paper sets out to inject task-related knowledge, e.g., conformational information, into PTPMs {to produce more informative protein representations.}
%the model input should not only include amino acid sequences but also task-desired signals to indicate what context the protein is in. 
%\textbf{Towards this end, we study how to inject task-desired signals to PTPMs to improve performance on the task of interest.} We achieve this goal by prompt learning.
%
Recently, prompts have been proposed to avoid fine-tuning PTLMs, which leads to improved performance. 
They are a sequence of discrete words designed by humans or continuous vectors learned through back-propagation.
Prompt learning aims to close the gap between pre-training and downstream tasks by converting the latter into the former. %A prompt is a word sequence designed by humans and later continuous  
{In this regard, prompts are supposed to contain task-related knowledge so as to induce PTLMs to make correct predictions~\cite{DBLP:journals/corr/abs-2103-08493}.  
Naturally, we are curious if we can leverage prompts to inject task-related knowledge into PTPMs. 
% to obtain task-aware protein representations.
However, humans still cannot completely understand the life language, i.e., the amino acid sequence. 
It is thus infeasible to design prompts based on the amino acid vocabulary.
}

To solve this issue, we attempt to learn new prompts, which are out of the amino acid vocabulary, from the task of interest. 
{The prompts can be plugged into PTPMs and optimized through back-propagation.
During the optimization process, PTPMs can acquire task-related knowledge and produce enhanced protein representations.
%to inject desired signals to PTPMs.
}
Taking conformational knowledge injection as an illustrative example, we develop a conformation-aware pre-trained protein model (ConfProtein). 
%Tailoring protein conformation, 
Specifically, ConfProtein has two learnable prompts for the properties of the protein itself and the interaction conformation in protein pairs respectively.
% the native conformation of single proteins and the interaction conformation in protein pairs respectively. 
As for the properties of the protein itself can be mined by the {sequence} of amino acids, we leverage the {masked language modeling (MLM)} task~\cite{devlin-etal-2019-bert} to learn this prompt, which is called the \emph{sequence prompt} (Seq prompt).
For conformations that exist in interaction pairs, an \emph{interaction conformation prompt} (IC prompt) is learned with protein-protein interaction prediction (PPI) tasks. Two prompts can be learned in a multitask setting.

We train the ConfProtein on the physical-only protein interaction network that contains 12,106 species. 
The experimental results show that PTPMs with the Seq prompt can only acquire knowledge about amino acid sequences and relevant secondary structures, while those with the IC prompt can effectively acquire 3D structural knowledge. 
Notably, resulting PTPMs with appropriate learned prompts outperform state-of-the-art (SOTA) models while inappropriate prompts will degrade their performance.
%which means the model can be a probe to find task-relevant conformation.
%Whether or not a conformational prompt is added, protein embeddings always contain non-conformational information such as secondary structure.
%
The main contributions of this paper are summarized as follows:
\begin{itemize}
\item We propose to learn {pluggable, interpretable and extensible prompts} to inject task-related knowledge into pre-trained protein models.% to obtain task-aware representations. 
\item As an instantiation, we design the ConfProtein model that injects sequential and conformational knowledge into pre-trained protein models in a multitask setting.
%, and designed the IC prompt to inject conformational knowledge.
%To the best of our knowledge, prompt engineering is a new branch of research that has not been explored in protein pre-trained models and training a prompt is also unexplored.
\item We created a new dataset that contains interaction conformational information for contact prediction.
\item A comprehensive evaluation on protein function and structure prediction tasks shows proper prompts significantly improve pre-trained models' performance.
%a context token contributes towards improvements its related tasks and training a new token do not make other token be catastrophic forgetting.
\end{itemize}

\section{Related Works}
\textbf{Pre-trained Protein Models} 
%Unlabeled data accounts for the majority of the overall data, and how to extract information from these data is critical to the effect of the model.
As PTLMs have been proved effective in natural language processing (NLP)~\cite{devlin-etal-2019-bert, NEURIPS2020_1457c0d6, DBLP:journals/jmlr/RaffelSRLNMZLL20}, some works try to extend such models to images~\cite{dosovitskiy2020image} and proteins.
\citeauthor{Rivese2016239118}~\yrcite{Rivese2016239118} firstly explore whether the Transformer architecture can be used to deal with proteins and find that the features learned by PTPMs contribute to the structure prediction performance.
\citeauthor{9477085}~\yrcite{9477085} conduct comprehensive experiments to study the limits of up-scaling PTPMs and show that PTPMs with a single protein input can obtain comparable performance to the top prediction methods in computational biology based on multiple sequence alignment (MSA).
To figure out why PTPMs work, \citeauthor{vig2021bertology}~\yrcite{vig2021bertology} focus on reconciling attention with known protein properties and identify that different layers can capture different structural information.

A growing body of works points out that general protein representations cannot meet the needs of describing specific properties in various biological processes.
To incorporate co-evolutionary signals from MSAs, \citeauthor{pmlr-v139-rao21a}~\yrcite{pmlr-v139-rao21a} develop the MSA Transformer and prove that homologous protein sequences can provide {native conformational} information and promote contact prediction performance. 
Moreover, MSAs inevitably contain non-homologous residues and gaps. \citeauthor{zhang2021coevolution}~\yrcite{zhang2021coevolution} introduce Co-evolution Transformer which considers the relationship between MSAs and the target protein and mitigates the influence of non-homologous information.
Both models mine co-evolutionary information from homologous protein sequences which have similar amino acid sequences and achieve auspicious performance.
However, using heterogeneous models to capture homologous protein patterns will greatly limit the application scenarios and it is hard to extend the models with other task-related knowledge.

\textbf{Prompts for Pre-trained Models} 
Prompts are introduced in GPT-3~\cite{NEURIPS2020_1457c0d6} and researchers have shown that prompts can be designed or learned to capture the task-related information~\cite{DBLP:journals/corr/abs-2107-13586}. 
\citeauthor{schick-schutze-2021-exploiting}~\yrcite{schick-schutze-2021-exploiting} introduce pattern-exploiting training (PET) and demonstrate that providing task descriptions to PTLMs can be comparable with standard supervised fine-tuning.
To avoid the disturbance of human bias, \citeauthor{gao2021making}~\yrcite{gao2021making} propose LM-BFF which utilizes generative models to obtain prompt templates and label tokens.
%Recently, many works have realized the limitations of finding prompts in discrete spaces,
%However, as neural networks are inherently continuous, searching prompt tokens over the discrete space $\mathcal{V}$ can lead to sub-optimal performance~\cite{DBLP:journals/corr/abs-2103-10385, DBLP:journals/corr/abs-2104-05240, DBLP:journals/corr/abs-2108-13161}. 
However, the above discrete prompt setting inherently requires the tokens in the vocabulary, which may limit the capability of prompt-based models.
%and the prompt distribution could be high influenced by the vocabulary distribution.
One solution is to find optimal prompt vectors in continuous spaces~\cite{DBLP:journals/corr/abs-2103-10385, DBLP:journals/corr/abs-2101-00190, DBLP:journals/corr/abs-2108-13161}.
Despite of promising performance~\cite{DBLP:journals/corr/abs-2101-00190, DBLP:journals/corr/abs-2104-08691}, these prompts are task-oriented, and thus suffers from low generality. 
Also, the learned continuous prompts are of low interpretability~\cite{hambardzumyan-etal-2021-warp}.
%To avoid engineering prompts and optimizing additional parameters, we aim to design a novel approach to obtain prompts for pre-trained protein models. Furthermore, the proposed approach is interpretable, can be plugged into any pre-train models and extended to multiple scenarios.

\section{Background}

\begin{figure}[t]
\begin{center}
\centerline{\includegraphics[width=0.95\columnwidth]{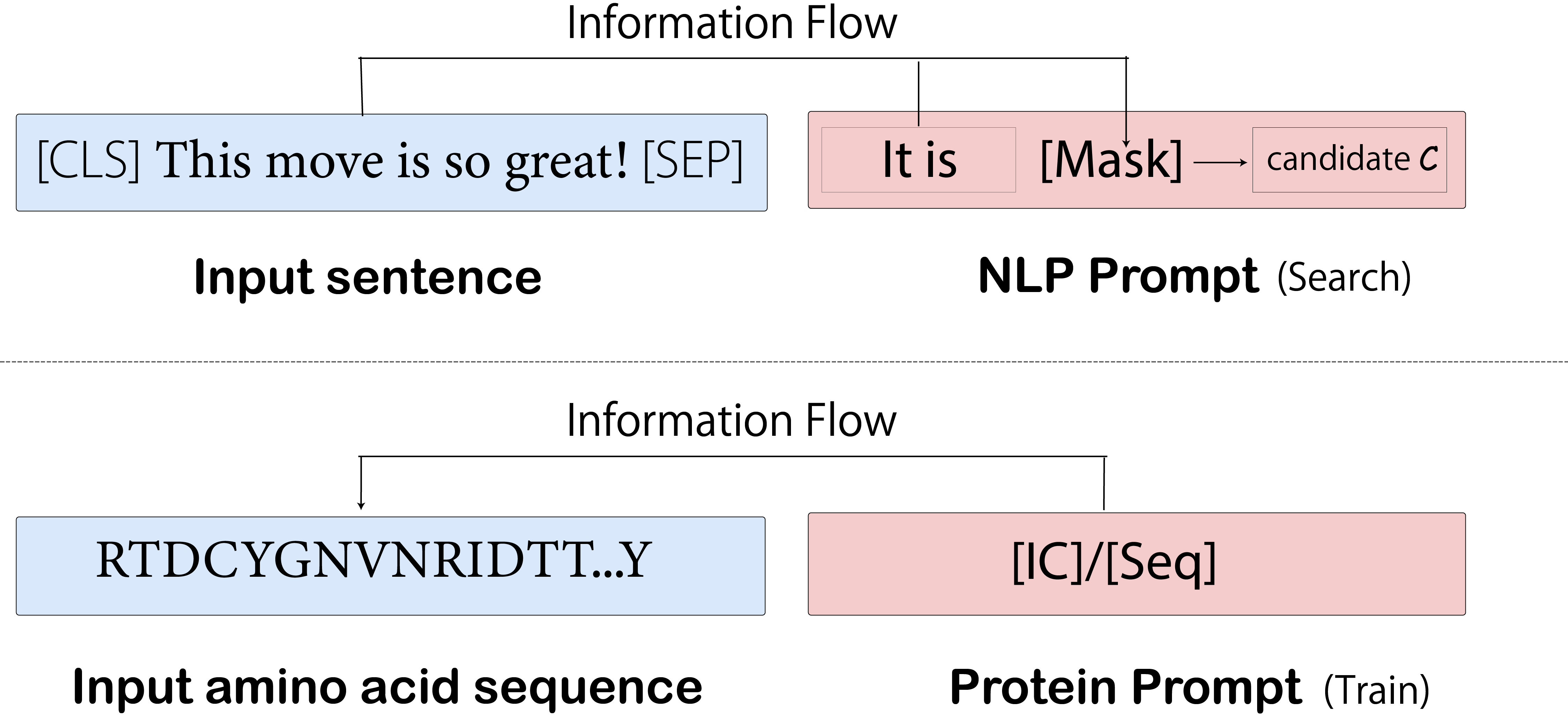}}
\caption{The relationship between NLP Prompts and Protein Prompts. \textbf{Top} (NLP Prompts): Prompt engineering aims to search a task-oriented pattern string (a template which contains $\mathtt{[MASK]}$) and a set of candidates in the embedding space. Given an input sentence, since $\mathtt{[MASK]}$ is a part of the pre-training MLM objective, its representation can be determined by the input and the template. \textbf{Bottom} (Protein Prompt): we aim to train a semantic token (such as $\mathtt{[IC]}$ which contains interaction conformational information). Given an amino acid sequence, the semantic token can be trained to provide protein representations with task-related information. }

\label{prompt-different}
\end{center}
\vskip -0.3in
\end{figure}

\subsection{Protein Conformation}
\label{biological}
It is well known that the structure of a protein determines its function.
Protein structures can be described from four levels: the primary structure (sequence of amino acids), the secondary structure (highly regular local sub-structures, such as $\alpha$-helix and $\beta$-sheet), the tertiary structure (three-dimensional structure), and the quaternary structure (complex). 
Given a sequence of amino acids, the primary structure and the secondary structure (except disordered regions) are determined, unless the protein is denatured.
Various 3D conformations are the major obstacle to predicting protein functions from its structure. 

The alternative structures of the same protein are referred to as different conformations, and transitions between them are called conformational changes. 
Since macromolecules are not rigid, protein structures can undergo reversible changes in response to various biological processes.
% Proteins can shift between several related conformations while they perform their biological functions.
The native conformation (NC) refers to the 3D structure into which a protein naturally folds, and the interaction conformation (IC) refers to the counterpart that a protein folds to interact with others when achieving its biological functions.
Although great breakthroughs have been made in the study of protein native conformation~\cite{alphafold2}, other conformations of proteins remain to be explored.
In this paper, we mainly focus on interaction conformation from which we can have a deeper understanding of PPI.

\subsection{Prompt Learning}
\label{prompt}
%\bm{x}_{\mathtt{in}} =
Given an input sequence $ S_{\mathtt{in}}=\{s_{\mathtt{in}}^1, s_{\mathtt{in}}^2, \ldots, s_{\mathtt{in}}^n\}$ where $s_{\mathtt{in}}^i \in \mathcal{V}$ is the $i$-th token in the sequence, $n$ is the sequence length and $\mathcal{V}$ is a vocabulary. With typical pre-trained models, an embedding operator is defined as
\begin{equation}
\label{general_emb}
E(\cdot) = E_{\mathtt{tok}}(\cdot) + E_{\mathtt{seg}}(\cdot) + E_{\mathtt{pos}}(\cdot),
\end{equation}
which is the sum of the corresponding token, segment, and position embeddings.
Thus we can have $E({s}_{\mathtt{in}}^i) = \bm{x}_{\mathtt{in}}^i$ and $E(S_{\mathtt{in}}) = \{\bm{x}_{\mathtt{in}}^1, \bm{x}_{\mathtt{in}}^2, \ldots, \bm{x}_{\mathtt{in}}^n\}=\bm{X}_{\mathtt{in}}$.
A pre-trained model $\mathcal{M}$ conducts a mapping $f: \mathcal{X} \rightarrow \mathcal{H}$, where $\mathcal{X}$ is the space of the input sequence embeddings and $\mathcal{H}$ is the space of the returned representations $\bm{h}=f(\bm{X}_{\mathtt{in}})$. 
It is $\bm{h}$ that represents the input sequence as a dense vector. 

Prompt learning refers to those methods that utilize prompts to improve pre-trained models.
Originating from the NLP area, {prompts are designed to contain task-related information, and plugging prompts to pre-trained models can make them aware of the task of interest.}
%induce the pre-trained language model to output the correct prediction for a given task.
Conventionally, a prompt%, denoted as $\bm{p}$, 
is a sequence of tokens in a vocabulary $\mathcal{V}$. 
Figure~\ref{prompt-different} shows an example of NLP prompts. The given task is to classify emotions of the input sequence \emph{This movie is so great!}.
%and a symbolized token $\bm{x}$ (e.g., $\mathtt{[MASK]}$). 
With the prompt (\emph{It is $\mathtt{[Mask]}$}), emotion classification is converted to the prediction of the masked word, i.e., \emph{good} for the positive emotion class and \emph{bad} for the negative emotion class.
%
%we can obtain the probability distribution of every class ($y\in Y$) as:
% \begin{equation}
% p(y|\bm{x}, \bm{p}) = \sum_{w\in\mathcal{C}_y}p(\mathtt{[Mask]}=w | \bm{x}, \bm{p}),
% \end{equation}
% where $\mathcal{C}_y$ is the set of candidate words corresponding to the emotion class $y$.
This is in line with the pre-training MLM objective.
%, where the task is to recover what is replaced by $\mathtt{[Mask]}$. 
Thus the gap between the pre-training and the downstream classification task is closed and the pre-trained model is able to achieve better performance.
%on the downstream task.  
%
% \begin{equation}
% \bm{x} || \bm{p} = \{\bm{x} || \mathtt{[SEP]} || \bm{\tau},\bm{x}\}.
% % \bm{x}_{\mathtt{prompt}} = \mathtt{[CLS]}\bm{x}_{\mathtt{in}}\mathtt{[SEP]}\tau_{\mathtt{task}}\mathtt{[MASK]}.
% \end{equation}
%
%As the $\mathtt{[Mask]}$ token is used along with the template \emph{It is} during training language models, we interpret that this token has acquires the ability to inject task information, i.e., cloze-comprehension, into PTLMs.
%
%
\begin{figure*}
%\vskip 0.2in
\begin{center}
\centerline{\includegraphics[width=0.65\textwidth]{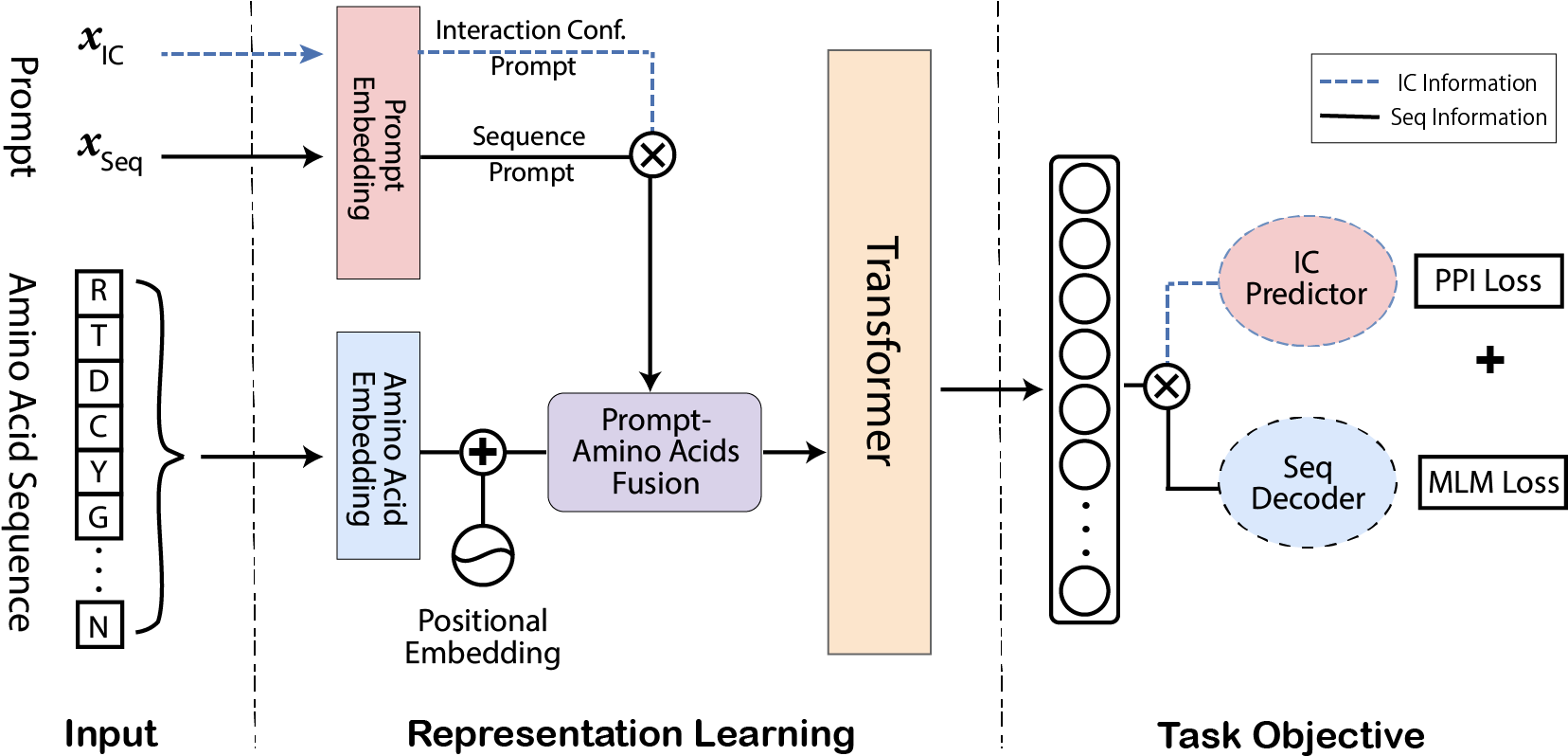}}
\caption{Overview of our proposed ConfProtein.}
\label{ConfProtein-graph}
\end{center}
\vskip -0.2in
\end{figure*}
The above example implements prompts as a task-oriented sequence 
%$\bm{p} =\{E(s'_{\mathtt{in}}^1),E(s'_{\mathtt{in}}^2),\ldots,E(s'_{\mathtt{in}}^m)\}$, where $s'_{\mathtt{in}}^j \in \mathcal{V}$ denotes the $j$-th 
with each discrete token from the vocabulary $\mathcal{V}$.
In contrast, the continuous prompt approach searches in a continuous embedding space $\mathcal{X}$.
%$\bm{p} = \{\bm{x}''_1,\bm{x}''_2,\ldots,\bm{x}''_m\}$, where $\bm{x}''_j \in \mathcal{X}$ is the $j$-th embedding used in continuous prompts. 
Note that each token is not limited to vectors converted from tokens in the vocabulary $\mathcal{V}$. 
%Because the prompt is searched by gradient-based methods, one can find the optimal continuous values for $\bm{p}$.
As humans cannot completely understand amino acid sequences, 
{it is infeasible to design discrete prompts based on the amino acid vocabulary while existing continuous prompts are task-oriented and suffers from low generalization.
}

\section{Method}

{We propose to learn pluggable, interpretable, and extensible protein prompts that enable PTPMs to produce more informative representations.}
We will first introduce how to learn such prompts in Section~\ref{learningprompts},
%the methodology of how to inject task-desired signals into pre-trained models by 
then instantiate a conformation-aware pre-trained protein model in Section~\ref{ConfProtein}. Figure~\ref{ConfProtein-graph} shows the overview of ConfProtein.

\subsection{
Protein Prompt Learning
}
\label{learningprompts}
We start this section by re-interpret the concept of prompts: {A prompt is a symbolized pattern string that can be manually designed or automatically learned to inject task-related knowledge to pre-trained models so as to produce informative representations.}
With prompt learning, the model input consists of two parts -- the original input sequence $S_{\mathtt{in}}$ and the prompt $S_{\mathtt{pt}}$. 
For the original input $S_{\mathtt{in}}$, we use the embedding operator defined in Equation~\ref{general_emb} to produce its embedding, i.e., $\bm{X}_{\mathtt{in}}=E(S_{\mathtt{in}})$. As for $S_{\mathtt{pt}}$, we assume the effects exerted by prompts on the input sequence $S_{\mathtt{in}}$ is not disturbed by the positions of prompts, % (that is the length of  $S_{\mathtt{in}}$),  %
%Since $s_p$ is not part of amino acid sequence and the effect of the prompt token on every amino acid should be indistinguishable, 
hence we do not add position and segment embeddings to prompt embeddings.
%Starting from scratch, we initialize a prompt as an one-hot vector $s_p$ that is later converted to be a dense vector: 
That is, $\bm{X}_{\mathtt{pt}} = E_{\mathtt{tok}}(S_{\mathtt{pt}})=\{E_{\mathtt{tok}}(s^1_{\mathtt{pt}}),\dots, E_{\mathtt{tok}}(s^m_{\mathtt{pt}})\}$. 
The whole model input can be denoted as:
\begin{equation}
    \bm{X}_{\mathtt{prompt}} = \bm{X}_{\mathtt{in}} || \bm{X}_{\mathtt{pt}},
\end{equation}
where $||$ denotes the concatenation operation between two vectors. The length of the whole sequence thus is $n+m$.

%Hence the learned prompt can go beyond the vocabulary $\mathcal{V}$. 

% \begin{definition}
% \label{def:protein prompt}
% A prompt is a set of symbols that can be manually designed or automatically learned to inject task-desired information to protein representation.
% \end{definition}
%
%
%Suppose the downstream task desire the information which corresponds to prompts $\{p_{\mathtt{IC}}, p_{\mathtt{MLM}}\}$. The input can be reformulated as:
% \begin{equation}
%     \bm{x}=\bm{x}_{\mathtt{in}}p_{\mathtt{IC}}p_{\mathtt{MLM}}
% \end{equation}
%
%
%We use $p_{\mathtt{IC}}$ and $p_{\mathtt{Seq}}$ to denotes the tokens which is called the prompt token. 
%

\begin{figure}[t]
%\vskip 0.2in
\begin{center}
\centerline{\includegraphics[width=0.6\columnwidth]{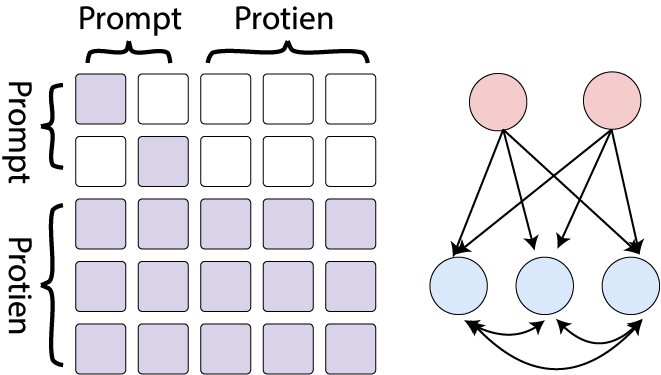}}
\caption{The knowledge-injection based attention masks. The original input tokens are denoted as blue circles while the prompt tokens are denoted as pink circles.}
\label{protein-prompt}
\end{center}
\vskip -0.3in
\end{figure}

With the self-attention mechanism in the Transformer architecture, each token in the whole sequence can attend to others at any position. However, {prompts are supposed to provide task-related information to the representation of the original input sequence}, so we only allow the one-way information flow from prompts to the original input, as illustrated in Figure~\ref{protein-prompt}. The information flow from the original token to the prompt tokens is thus forbidden. Also, to promote orthogonality and generalization, information flows between prompt tokens are forbidden.
%Inspired by~\cite{10.5555/3454287.3454804}, 
We design an attention mask matrix $M$ to fulfill this need. Let $M_{ij}$ denote  the $(i,j)$-element of the mask matrix, and we define:
\begin{equation}
    M_{ij} = \left\{
    \begin{aligned}
         0&, ( 1\leq i \leq m \mathtt{\ and\ } m < j \leq {m+n}) \mathtt{\ or}\\
          &\;\; (1 \leq i, j \leq m \mathtt{\ and\ } i \neq j)\\
         1&, \mathtt{others},
    \end{aligned}
    \right.
\end{equation}
%propose a novel knowledge-injection based attention masks. 
%
%Denote $M$ as the attention masks and $H$ as the hidden representation from the last attention layer, 
then the output calculation is modified as:
\begin{equation}
    \bm{h} = g(\mathtt{softmax}(\frac{QK^T}{\sqrt{d}}) \cdot M \cdot V),
\end{equation}
where $Q$, $K$, and $V$ are the linear projection of the token embedding $\bm{x}_{\mathtt{in}}$ and $\bm{x}_{\mathtt{pt}}$, $d$ is the hidden dimension, and $g(\cdot)$ denotes the other operations on top of self-attention in the Transformer architecture, such as skip connections and feed-forward networks.
After stacking multiple self-attention and other operations, the final representation, still can be denoted as $\bm{h}$, can contain the information from both the original sequence and the prompts. 
%\newpage

The loss function of {protein prompt learning} has two-folds:
%The information from untrained tokens is lack of interpretability. 
%To define the semantics of prompt token, the pre-training of prompt has two objectives: 
the knowledge conservation objective $\mathcal{L}_{C}$ and knowledge injection objective $\mathcal{L}_{P}$. The former tries to make pre-trained models preserve what has been learned from pre-training tasks while the latter aims to guide pre-trained models to acquire new knowledge.

{\textbf{Knowledge Conservation Objective}} In order to preserve the knowledge in the previously trained model, we calculate the previous task loss based on the returned {representation}:
\begin{equation}
    \mathcal{L}_{C}=\mathcal{L}_\mathtt{pr}(\bm{h}),
\end{equation}
where $\mathcal{L}_\mathtt{pr}$ is the loss function of the previous task, e.g., the MLM loss function during pre-training.

%suppose $\mathcal{L}_{p}$ be the loss related to prompt token $\bm{p}$. 

{\textbf{Knowledge Injection Objective}} As we expect the pre-trained model to grasp new knowledge by learning prompts, we calculate the loss on tasks where the knowledge of interest exists. %$p_\mathtt{new}$.
%can provide protein representations with a special information. 
%Since macromolecules are a complex system, every data point is a superposition of many pieces of information. What we can do is take advantage of the diversity of the data to get as orthogonal information as possible. 
We assume a specific type of knowledge, such as protein conformations, can be learned from multiple relevant tasks, such as protein-protein interaction and binding affinity prediction.
%Suppose a collection of tasks $T$ contain a task $t \in T$ which are related to the information. 
Let $\mathcal{L}_{\tau}$ be the loss of the task $\tau$ from the relevant task collection $\mathcal{T}$, such as the PPI prediction loss. We denote this loss function as following:
\begin{equation}
    \mathcal{L}_{I} = \sum_{\tau\in \mathcal{T}}\alpha_{\tau}\mathcal{L}_\tau(\bm{h}) 
\end{equation}
By optimizing $\mathcal{L}_{I}$, the pre-trained model can produce an informative representation {based on the knowledge associated with the tasks in $\mathcal{T}$.} Generally, we hope the pre-trained model not only preserves already learned knowledge but also grasps new skills, hence we have the following training object in a multitask setting:
\begin{equation}
    \mathcal{L} = \mathcal{L}_C + \lambda\mathcal{L}_{{I}}
\end{equation}
where $\lambda$ is the hyper-parameter balancing the previous and new losses.

% Different from prior works, we choose to train a new token as a protein prompt. We denote a protein as $P_{\mathtt{in}} = \{p_1, p_2, \ldots, p_n\}$ where $p_i$ is the $i$-th amino acid and $n$ is the number of amino acid. Given a task, we need find a trained task-related conformation prompt $\tau_{\mathtt{conf}}$ and reformulate the input as:
% $$P_{\mathtt{prompt}} = \mathtt{[CLS]}P_{\mathtt{in}}\mathtt{[SEP]}\mathtt{\tau_{\mathtt{conf}}}$$
% When the input is fed into protein pre-trained model, we can obtain the protein embeddings with conformational information, which is crucial for downstream tasks. 

\subsection{ConfProtein}
\label{ConfProtein}
We now instantiate the proposed prompt learning method by ConfProtein.
%enabling a PTPM to capture protein conformations.
%pre-trained with the MLM objective
We use the previous pre-training MLM task, which is to recover the replaced amino acids given the context, to optimize the PTPM and the Seq prompt $\boldsymbol{x}_{Seq}$. 
%To maintain the ability of protein understanding, we utilize MLM objective to guarantee model behavior with only amino acids as input. 
%As illustrated in Figure~\ref{model}, if there is no conformational prompt, 15\% amino acids will be randomly masked, and the masked amino acids prediction will be conducted. 
%
%We denote the masked amino acid sequence as $S_{\mathtt{masked}}$, and 
Let $Y$ be the set of masked out tokens, the MLM loss can be formulated:
% \begin{equation}
%     q(y_i|S_{\mathtt{masked}}) = {\frac{\exp(p(y_i|S_{\mathtt{masked}}))}{\sum_{v\in\mathcal{V}}\exp(p(v|S_{\mathtt{masked}}))}},
% \end{equation}
% \begin{equation}
%     \mathcal{L}_{C} = \sum_{y_i \in Y}-\log{q(y_i|S_{\mathtt{masked}})}.
% \end{equation}
\begin{equation}
    q(y|\bm{h}) = {\frac{\exp(p(y|\bm{h}))}{\sum_{v\in\mathcal{V}}\exp(p(v|\bm{h}))}},
\end{equation}
\begin{equation}
    \mathcal{L}_{C} = \sum_{y \in Y}-\log{q(y|\bm{h})}.
\end{equation}
The resulting representation should capture the chemical properties between amino acids and contributes to the prediction of protein secondary structures.
Further, we aim to inject protein conformational knowledge into the PTPM by learning the interaction-conformation prompt $\boldsymbol{x}_{IC}$.
%To construct IC prompt token, we use the mean of amino acid embeddings to represent the protein, and use a linear classifier $L$ to 
Towards this end, we conduct the new task -- predict whether the $p$-th and $q$-th proteins can interact with each other. 
%Let $\bm{x}^n=\{\bm{x}_1, \bm{x}_2, \ldots, \bm{x}_n\}$ be the $n$ proteins in subgraph, and $Y\in \{0, 1\}^{n\times n}$ is the interaction map.
The loss of the PPI task is as follows:
\begin{equation}
    \mathcal{L}_{I}(\bm{h}_p, \bm{h}_q) = 
    %\sum_{\mathbf{x}_i\in \mathbf{x}^n}\sum_{\mathbf{x}_j\in \mathbf{x}^n}
    \mathtt{BCE}(p(y_{p,q} | \bm{h}_p, \bm{h}_q)),
\end{equation}
where BCE is the binary cross-entropy loss function. 
% Then we have the following training object:
% \begin{equation}
%     \mathcal{L}=\mathcal{L}_{\mathtt{MLM}} + \lambda \mathcal{L}_{\mathtt{prompt}}(\mathtt{[CA]})
% \end{equation}

% By optimizing $\mathcal{L}_S$, the protein model can maintain its ability to gather information form sequence.

% \textbf{knowledge injection Objective} The knowledge injection objective aims to mine information from external database and inject it to prompt. As shown in Figure~\ref{model}, if a conformation prompt added on protein, we will use protein embedding to train on conformation related tasks ($T$), the loss is as follows:
% $$\mathcal{L}_{\mathtt{conf}} = \sum_{t\in T} \mathcal{L}_t$$
% where $\mathcal{L}_t$ is a specific task loss, and the conformation loss $\mathcal{L}_{\mathtt{conf}}$ represents the conformation knowledge injection loss. Then we have the following training objective:
% $$\mathcal{L} = \mathcal{L}_S + \lambda\mathcal{L}_{\mathtt{conf}}$$
% where $\lambda$ is the hyper-parameter.

\begin{table*}[t]
\caption{Results on Protein-Protein Interaction Prediction Tasks. There are two types of models that we compare with. The first four baselines are non pre-trained models including CNN, RCNN, LSTM, and GNN. The other baselines are pre-trained ones. For fair comparison, we only use pre-trained models to generate amino acid embeddings and feed these embeddings into GNN-PPI. Note that we do not modify the hyperparameters of GNN-PPI. The reported results are mean(std) micro-averaged F1 score.}
\label{ppi-experiment}
\footnotesize
\begin{center}
\begin{small}
\begin{sc}
\resizebox{0.78\textwidth}{!}{
\begin{tabular}{lccccccc}
\toprule
\multirow{2}{*}{Method}& 
\multicolumn{2}{c}{\textbf{SHS27k}} &
\multicolumn{2}{c}{\textbf{SHS148k}} &
\multicolumn{2}{c}{\textbf{STRING-HomoSapiens}} \\
& BFS & DFS & BFS & DFS & BFS & DFS\\
\midrule
DPPI & 41.43(0.6) & 46.12(3.0) & 52.12(8.7) & 52.03(1.2) & 56.68(1.0) & 66.82(0.3)\\
DNN-PPI & 48.09(7.2) & 54.34(1.3) & 57.40(9.1) & 58.42(2.1)  & 53.05(0.8) & 64.94(0.9)\\
PIPR & 44.48(4.4) & 57.80(3.2) & 61.83(10.2) & 63.98(0.8)  & 55.65(1.6) & 67.45(0.3)\\
GNN-PPI & 63.81(1.8) & 74.72(5.3) & 71.37(5.3) & 82.67(0.9) & 78.37(5.4) & 91.07(0.6)\\
\midrule
ProtBert & 70.94 & 73.36 & 70.32 & 78.86  & 67.61 & 87.44\\
OntoProtein & 70.16 & 76.35 & 67.66 & 77.56  & 70.59 & 81.94\\
ESM-1b & 68.12(1.9) & 75.80(2.4)& 68.74(1.4) & 75.16(2.8) & 76.85(0.7) & 86.66(0.1)\\
\midrule
ConfProtein-w/o-IC & 68.49(4.1) & 75.64(3.7) & 68.90(2.7) & 74.29(2.8) & 77.17(0.9) & 86.67(0.3) \\
ConfProtein & \textbf{71.24(3.5)} & \textbf{77.62(2.6)} & \textbf{72.23(3.4)} & 79.55(1.7)  & \textbf{78.26(0.2)} & 87.82(0.4)\\
\bottomrule
\end{tabular}
}
\end{sc}
\end{small}
\end{center}
\vskip -0.15in
\end{table*}

\section{Experiments}
\label{experiment}
%This section describes the experimental setup.
%\subsection{Experimental Setting}

\textbf{Pre-training Dataset}
We use the STRING dataset~\cite{doi:10.1093/nar/gky1131} that contains protein-protein interaction pairs for model pre-training.
Some interactions in the STRING dataset do not form stable conformations. 
To remove unstable conformations, we choose the physical-only interaction subset from STRING. The subset contains 65 million protein sequences from 14,094 species and 2.7 billion protein-protein interaction pairs. 
A PPI network can be defined, in which a node represents a protein and an edge represents the two interacting proteins.
The edge between protein pairs indicates that there is evidence of their binding or forming a physical complex. 
%
%
%Note that there are no interactions between two proteins from different species.
%
%
Similar to previous works, we reserve the Homo sapiens (a species contained in STRING) PPI pairs for downstream evaluation.

\textbf{Downstream Datasets}
Researchers use the PPI data of the Homo sapiens to create downstream datasets. \citeauthor{10.1093/bioinformatics/btz328}\yrcite{10.1093/bioinformatics/btz328}) created the SHS27k and SHS148k datasets based on {a random selection of} the Homo sapiens PPI data, and~\citeauthor{DBLP:journals/corr/abs-2105-06709}\yrcite{DBLP:journals/corr/abs-2105-06709} uses them all, which we denote as STRING-HomoSapiens. 
We leverage the PPI prediction task to evaluate whether the IC prompt can inject conformational knowledge into PTPMs.  Following~\citeauthor{DBLP:journals/corr/abs-2105-06709}\yrcite{DBLP:journals/corr/abs-2105-06709}, we regard PPI prediction as a link prediction task in the protein network. Two methods, i.e., Breath-First Search (BFS) and Depth-First Search (DFS), are used to split training and evaluation datasets for SHS27k, SHS148k, and STRING-HomoSapiens, respectively. 
Note that during pre-training, models are optimized to predict whether two proteins can interact with each other. While for downstream tasks, besides interaction prediction, models are required to predict interaction types.
%, which is much harder than the pre-training task.
We use the above three datasets for comparison with SOTA models in Section~\ref{main-results}, the SAbDab~\cite{10.1093/nar/gkt1043} dataset for the ablation study in Section~\ref{ablation-study}, and the TAPE~\cite{tape2019} dataset to analyze our model in Section~\ref{model-analyze}.
%Besides the PPI task in the STRING-HomoSapiens dataset,
The SAbDab~\cite{10.1093/nar/gkt1043} dataset is used for the prediction task of antibody-antigen binding affinity, which also requires protein conformational knowledge.
TAPE is a benchmark designed to evaluate the generalization of protein models. There are three major aspects that the benchmark involves: structure prediction, detection of remote homologs, and protein engineering. With TAPE, we can analyze and discuss the learned protein prompts.
%by answering the following questions.
Please see Appendix for the details of dataset statistics. 

\textbf{Pre-training} 
We implement the proposed ConfProtein using Pytorch~\cite{10.5555/3454287.3455008} and Fairseq~\cite{ott2019fairseq}. ConfProtein has 650M parameters with 33 layers and 20 attention heads. The embedding size is 1280. The batch size is set to be 20 and the learning rate is ${1\times10^{-5}}$ without weight decay. We use a fixed learning rate schedule. Limited by memory, at each step we randomly sample a small set of proteins from one species and the maximum number of amino acids is 2048.
All models are trained on 2$\times$A100 GPUs for 100k steps of updates. 
Unless otherwise specified, we use this model in all downstream experiments. The source code was uploaded and will be available online.

\textbf{Baseline} 
DPPI~\cite{10.1093/bioinformatics/bty573}, DNN-PPI~\cite{doi:10.3390/molecules23081923}, and PIPR~\cite{10.1093/bioinformatics/btz328}, and GNN-PPI~\cite{10.1093/bioinformatics/btz328} are not pre-trained models. The first three baselines use different deep learning architectures (CNN, RCNN, and LSTM) to convert amino acid embeddings to protein embeddings, and use linear classifiers to predict whether two proteins have an interaction relationship. GNN-PPI leverages graph neural networks to focus on the entire interaction graph and achieves SOTA performance.
ProtBert~\cite{9477085}, OntoProtein~\footnote{\url{https://openreview.net/forum?id=yfe1VMYAXa4}}, and ESM-1b~\cite{rao2021transformer} are three pre-trained models. ConfProtein-w/o-IC is a variant of our ConfProtein that {includes the Seq prompt and excludes the IC prompt}.

Note that OntoProtein incorporates information from knowledge graphs into protein representations. Since ConfProtein's architecture is the same as ESM-1b, 
%we illustrate the effect of the IC prompt on protein representations by comparing with ESM-1b and ConfProtein-w/o-IC.
{we examine if the Seq prompt can preserve sequential amino acid information by comparing ConfProtein-w/o-IC and ESM-1b, and if the IC prompt can inject conformational knowledge by comparing ConfProtein and ConfProtein-w/o-IC}.
For fair comparison with PTPM baselines, following {OntoProtein}, we freeze the parameters of PTPMs and utilize GNN-PPI to predict interaction types.
%They encode amino acid sequences into the protein embeddings that serve as the initialization of nodes in GNN-PPI.
%, which is then input to the GNN-PPI model to predict interaction relationships.
%replace the initial protein embedding part of GNN-PPI with ProtBERT and OntoProtein as baselines
%we use pre-trained baselines as protein embedding encoder and use GNN-PPI to predict the relationship between two proteins. 
%Note that we do not modify the hyperparameters of GNN-PPI. 
%Since the limitation of memory, we use the mean of amino acid sequence representations as protein representations in STRING-HomoSapiens.

\section{Results and Discussion}
\subsection{Main Results}
\label{main-results}
%\textbf{Result} 

We present the evaluation results of the proposed ConfProtein and state-of-the-art baselines in Table~\ref{ppi-experiment}. 
%From Table~\ref{ppi-experiment}, 
By comparing with non-pre-trained baselines, we find that our proposed ConfProtein is better than DPPI, DNN-PPI, and PIPR on all datasets, and outperforms GNN-PPI on the small SHS27k dataset, 
which indicates the learned IC prompt improves PTPMs for PPI prediction. 
On the largest STRING-HomoSapiens dataset, the performance of GNN-PPI surpasses all the PTPMs. 
This is that, due to memory limitation, we calculate the mean of amino acid embeddings in a sequence as the protein representation. It is not as capable as the convolutional network in GNN-PPI. For small datasets including SHS27k and SHS148k, we do not have the memory issue, hence we employ the same convolutional network in PTPMs and the performance is better than GNN-PPI. 

\begin{figure}[t]
%\vskip 0.2in
\begin{center}
\centerline{\includegraphics[width=0.85\columnwidth]{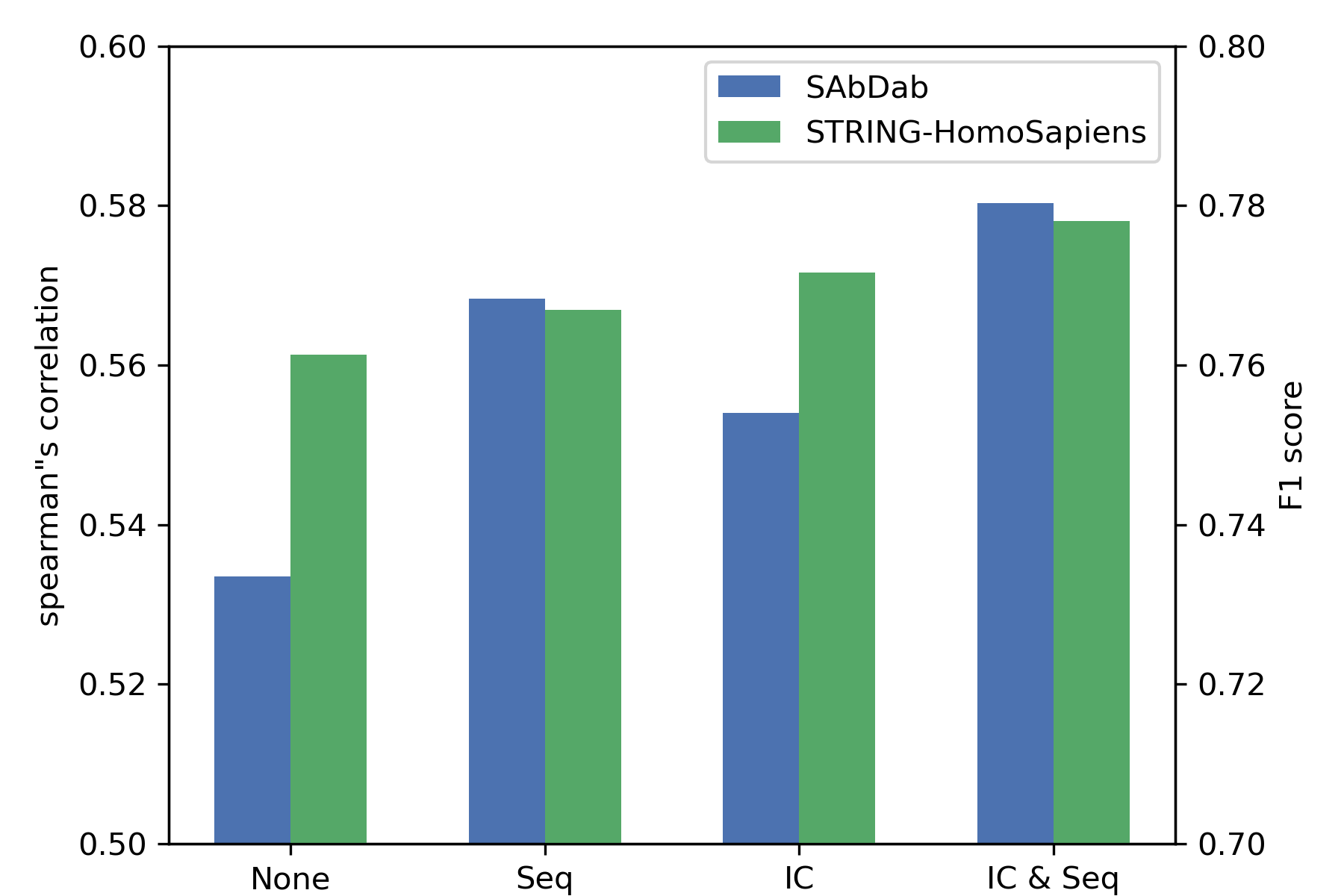}}
\caption{Ablation of ConfProtein with different prompts on SHS27k (F1 score) and SAbDab (spearman's $\rho$ ).}
\label{fig:ablation}
\end{center}
\vskip -0.4in
\end{figure}

\begin{figure*}
\begin{center}
\centerline{\includegraphics[width=0.9\textwidth]{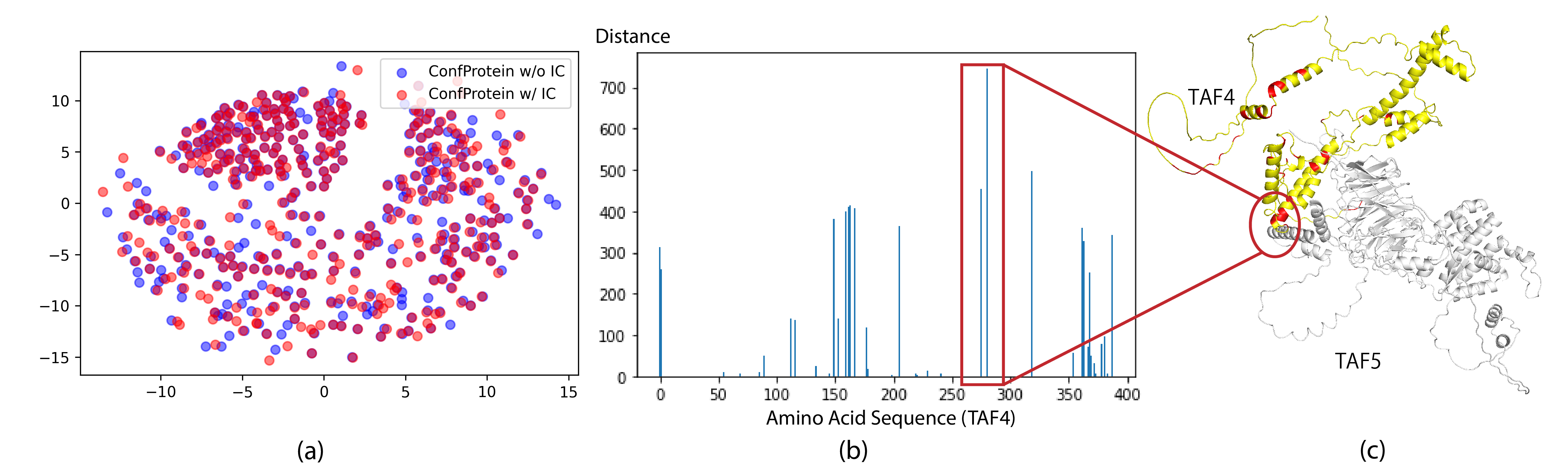}}
\vskip -0.1in
\caption{Visualization of interaction conformational information. The two proteins are Transcription initiation factor TFIID subunit4 (TAF4) and Transcription initiation factor TFIID subunit 5 (TAF5).\cite{doi:10.1126/science.abm4805}. \textbf{Left}: Visualize the embedding of amino acids (TAF4) with and without conformational information by MDS. \textbf{Middle}: Visualize distances of corresponding amino acids with and without conformational information. \textbf{Right}: Visualize amino acids with distances greater than 100 (red). }
\label{fig:case-study}
\end{center}
\vskip -0.3in
\end{figure*}

OntoProtein tries to make use of all the information in a knowledge graph. Our model is better than OntoProtein on all datasets, indicating that not all information in knowledge graphs has a contributing effect on PPI tasks and sometimes incorporating inappropriate knowledge can be harmful.
By comparing ESM-1b and ConfProtein-w/o-IC, we find that with {the knowledge conservation objective}, our model is still able to capture amino acid sequences information. 
Finally, the performance gap between the ConfProtein-w/o-IC and ConfProtein demonstrates that the IC prompt can indeed enhance the protein representations with interaction conformation knowledge.

\subsection{Ablation Study}
\label{ablation-study}
%The MLM task is to leverage the sequence information to predict masked amino acids. According to the MLM objective, $\mathtt{[CLS]}$ and $\mathtt{[SEP]}$ can capture the protein sequence information. By the mean time, we also use PPI task to make $\mathtt{[IC]}$ to provide protein with conformation information. Here, 
We conduct an ablation study on the SHS27k and SAbDab datasets to analyze the influence of the Seq and IC prompts.
From Figure~\ref{fig:ablation}, we observe that the IC prompt can improve our model's performance on the tasks of both PPI prediction and antibody-antigen binding affinity prediction, proving that prompts with explicit semantics can generalize to relevant tasks.
We also notice that the performance of ConfProtein degrades when the Seq or IC prompt is absent, demonstrating that these two prompts can conjunctionally inject task-related knowledge to pre-trained protein models. 
Furthermore, we notice that the Seq prompt has more influence in the SAbDab dataset, while in the STRING-HomoSapiens dataset, the IC prompt is more influential.
This is most likely because PPI is determined by protein conformations and {ConfProtein can acquire conformational knowledge via the IC prompt}. 
For the prediction of antibody-antigen binding affinity, the properties of amino acids play a key factor, which can be obtained by the Seq prompt. It also shows the learned prompts are extensible from PPI prediction to antibody-antigen binding affinity prediction.

\subsection{Analysis of the Learned Prompts}
\label{model-analyze}

\textbf{How can we interpret the IC prompt?}
Since the IC prompt is trained to provide PTPMs with conformational knowledge, we analyze what exactly the amino acid representations have changed. As shown in Figure~\ref{fig:case-study}(a), we firstly visualize the embeddings of amino acids of the TAF4 protein with and without the IC prompt based on multi-dimensional scaling (MDS)~\cite{mds}. Then we calculate the distances between {two embeddings of one amino acid} and plot them in Figure~\ref{fig:case-study}(b). We mark the embedding pairs with distances larger than 100 in red in Figure~\ref{fig:case-study}(c). We observe that the marked embeddings are all amino acids on the protein surface, which is consistent with the fact that the amino acids related to PPI are almost located on the surface of the protein, not the core~\cite{https://doi.org/10.1007/s10930-007-9108-x}.

\begin{table}[t]
\caption{Result on Contact Prediction Task}
%\vskip 0.15in
\begin{center}
\begin{small}
\begin{sc}
\begin{tabular}{lc}
\toprule
Method & CASP12 (P@L/2)\\
\midrule
ProtBert  & 0.35 \\
MSA Transformer & \textbf{0.49} \\
ESM-1b & 0.42 \\
\midrule
ConfProtein (w/o IC) & 0.43 \\
ConfProtein & 0.41 \\
\bottomrule
\end{tabular}
\end{sc}
\end{small}
\end{center}
\label{contact-experiment}
\vskip -0.3in
\end{table}

\textbf{Prompts can be negative to downstream tasks.}
The protein contact map represents the distance between all possible amino acid residue pairs.
The task of contact prediction is to classify whether a pair of amino acids contact, and can be used to evaluate the ability of PTPMs to capture conformational information. 
In TAPE, the CASP12~\cite{doi:10.1002/prot.25415} dataset is used to evaluate a model's performance on contact prediction. 
The 3D coordinates of atoms in CASP12 are experimentally measured when the protein naturally folds, hence the contact map corresponds to the native conformation.
To obtain the contact map corresponding to the interaction conformation, we build a protein structure dataset from~\citeauthor{doi:10.1126/science.abm4805}\yrcite{doi:10.1126/science.abm4805} that obtain through experiments the three-dimensional structures of proteins when they are interacting with others. 
The dataset, called ICProtein~\footnote{ ICProtein is available at \url{shorturl.at/sA345}}, contains 1,106 protein complexes, thus there are 2,212 contact maps. Details are in Appendix.

\begin{figure}
\begin{center}
\centerline{\includegraphics[width=0.95\columnwidth]{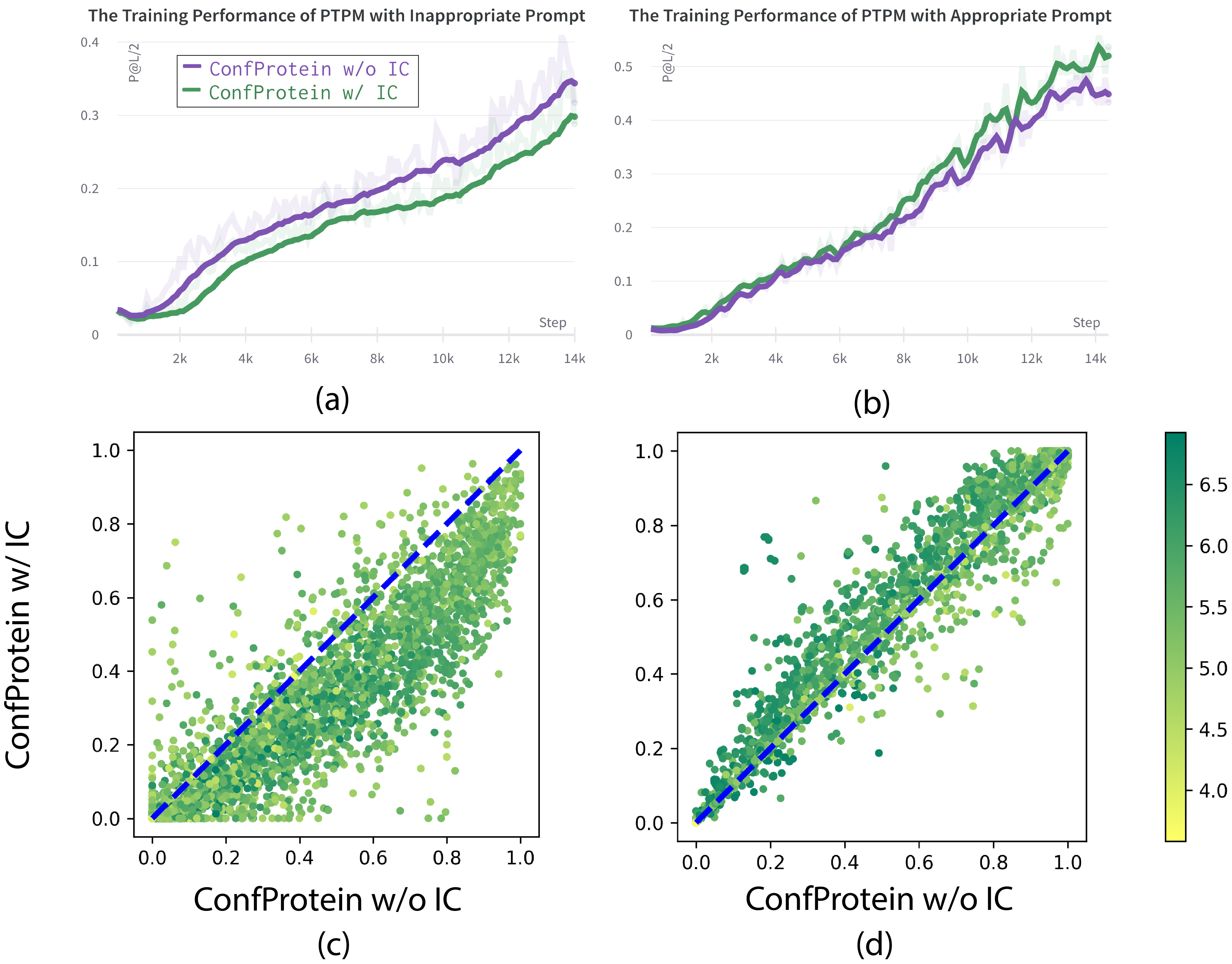}}
\caption{Comparison of adding inappropriate and appropriate prompts. \textbf{Left~(a,c)} : Use the IC prompt to predict NC contact map. \textbf{Right~(b,d)} : Use the IC prompt to predict IC contact map. \textbf{Up~(a,b)} : The training performance of PTPMs with different prompts on contact prediction tasks. \textbf{Bottom~(c,d)}: The long rang P@L/2 performance distribution of PTPMs on contact prediction. 
%Each data point is colored by the log of the number of amino acids in the protein.
}
\label{fig:contact-graph}
\end{center}
\vskip -0.45in
\end{figure}

We compare our model to two PTPMs (ProtBert and ESM-1b) which only leverage sequence information and the main differences between the two models are the number of parameters and their pre-training datasets.
We also compare to the MSA Transformer which interleaves row and column attention across the input aligned sequences.
We fit a linear classifier to predict whether two residues contact. We report the precision for the top $L/2$ contacts for medium- and long-range contacts, where $L$ is the length of the protein.

In Table~\ref{contact-experiment}, there is a large gap between the MSA Transformer and others. We attribute this gap to a strong correlation between {contact maps and the homologous information}. 
We notice that the IC prompt decreases performance, which means the model with an inappropriate prompt performs poorer than the model without any additional prompts.
As shown in Figure~\ref{fig:contact-graph}, the same prompt has diametrically opposite effects on the same task in different datasets, demonstrating that the use of inappropriate knowledge will produce negative impacts. We also notice that ConfProtein is more effective in long proteins. {This is because long protein sequences have greater differences between the contact maps of different conformations, and more 3D information is needed to accurately predict contact maps.}

\textbf{Prompts can be neutral to downstream tasks.} We leverage the secondary structure prediction task to explore how prompts perform on tasks that do not require them. The secondary structure is determined by patterns of hydrogen bonds between backbone amine and carboxyl groups. This means it is the amino acid sequence that determines the secondary structure. 
We also compare to ESM-1b, ProtBert, and MSA Transformer to figure out whether the IC prompt is helpful for the secondary structure prediction task. We report both 3-class and 8-class accuracy on a per-amino acid basis on the CB513~\cite{https://doi.org/10.1002/(sici)1097-0134(19990301)34:4<508::aid-prot10>3.0.co} dataset.

As analyzed before, the MSA Transformer can leverage MSAs to obtain native conformations, and the IC prompt can provide protein representations with interaction conformational knowledge. According to Table~\ref{ss-experiment}, we observe that almost every PTPMs achieve similar performance, indicating that conformational knowledge does not affect the performance on the secondary structure prediction task.

\begin{table}[t]
\caption{Results on Secondary Structure Prediction Tasks}
\label{ss-experiment}
%\vskip 0.15in
\begin{center}
\begin{small}
\begin{sc}
\begin{tabular}{lcc}
\toprule
Method & CB513-Q3 & CB513-Q8\\
\midrule
ProtBert  &  0.81 & 0.67 \\
MSA Transformer & - & 0.73 \\
ESM-1b & 0.84 & 0.71 \\
\midrule
ConfProtein (w/o IC) & 0.842 & 0.713 \\
ConfProtein & 0.836 & 0.707 \\
\bottomrule
\end{tabular}
\end{sc}
\end{small}
\end{center}
\vskip -0.2in
\end{table}

\begin{table}[t]
\caption{Results on Fluorescence \& Stability Tasks}
\label{fluorescence-experiment}
%\vskip 0.15in
\begin{center}
\begin{small}
\begin{sc}
\begin{tabular}{lcc}
\toprule
Method & Fluorescence&stability\\
\midrule
ConfProtein (w/o IC) &0.678 & 0.797 \\
ConfProtein & 0.660 &  0.793 \\
\bottomrule
\end{tabular}
\end{sc}
\end{small}
\end{center}
\vskip -0.2in
\end{table}

\textbf{Prompts can be knowledge probe for unknown downstream tasks.}
Since each prompt can be assigned specific semantics after learning, we utilize prompts as a knowledge probe to determine the information needed for each task. In this part, we will focus on two protein engineering tasks: fluorescence landscape prediction and stability landscape prediction.
The green fluorescent protein exhibits bright green fluorescence when exposed to light~\cite{doi:10.1021/bi00610a004}. The fluorescence landscape prediction task aims to map proteins to a log-fluorescence intensity. The protein stability is measured by the free energy difference between the folded and unfolded protein states. The stability landscape prediction task aims to map a protein to the label indicating the most extreme circumstances in which the protein can maintain its fold. Performance on these two tasks is measured by spearman's $\rho$ on the test set.

From the result in Table~\ref{fluorescence-experiment}, we can find that the performance of ConfProtein-w/o-IC is better than the performance of ConfProtein in the fluorescence task. 
From the analysis of the model on the CASP12 dataset in TAPE, we conclude that the information required for the log-fluorescence of proteins is incompatible with the IC prompt.
%, and this information is most likely some kind of 3D information.
This conclusion is consistent with the fluorescence mechanism that the process involves base-mediated cyclization followed by dehydration and oxidation~\cite{doi:10.1126/science.273.5280.1392}.
For the stability task, since the performance of ConfProtein-w/o-IC and ConfProtein is comparable, similar to the CB513 results, we believe that conformational knowledge has no effect on protein stability.

\section{Conclusion}
In this paper, we transfer the concept of prompts from NLP to protein representations. We present the conformation-aware pre-trained protein model with the sequence and interaction conformation prompts in a multi-task setting. People can leverage these prompts to achieve diverse protein representation. Experimental results on widespread protein tasks demonstrate that an appropriate prompt can provide task-related knowledge for protein representations. 
%The learned prompts can be combined and extended for new complex tasks.
%
%These positive results point to future work in (1) using the homologous protein database to provide models with native conformation prompt token; (2) exploring how to inject multiple knowledge into protein representations with minimal loss.

% In the unusual situation where you want a paper to appear in the
% references without citing it in the main text, use \nocite
\nocite{langley00}
\bibliography{main}
% \bibliography{example_paper}
\bibliographystyle{icml2022}

%%%%%%%%%%%%%%%%%%%%%%%%%%%%%%%%%%%%%%%%%%%%%%%%%%%%%%%%%%%%%%%%%%%%%%%%%%%%%%%
%%%%%%%%%%%%%%%%%%%%%%%%%%%%%%%%%%%%%%%%%%%%%%%%%%%%%%%%%%%%%%%%%%%%%%%%%%%%%%%
% APPENDIX
%%%%%%%%%%%%%%%%%%%%%%%%%%%%%%%%%%%%%%%%%%%%%%%%%%%%%%%%%%%%%%%%%%%%%%%%%%%%%%%
%%%%%%%%%%%%%%%%%%%%%%%%%%%%%%%%%%%%%%%%%%%%%%%%%%%%%%%%%%%%%%%%%%%%%%%%%%%%%%%
\newpage

\clearpage

\newpage
\appendix

\section{Construction of Pre-training Dataset}
To inject interaction conformation knowledge into the ConfProtein, we construct a PPI dataset -- a large-scale physical-only interaction network. We use the latest STRING database with only the physical-only mode, which means edges between the protein pairs indicate evidence of their binding or forming a physical complex. The database contains in total 65 million protein sequences from 14,094 species and 2.7 billion protein-protein interaction pairs. Note that there is no edge between proteins that come from different species.

\begin{figure}[h]
%\vskip 0.2in
\begin{center}
\centerline{\includegraphics[width=\columnwidth]{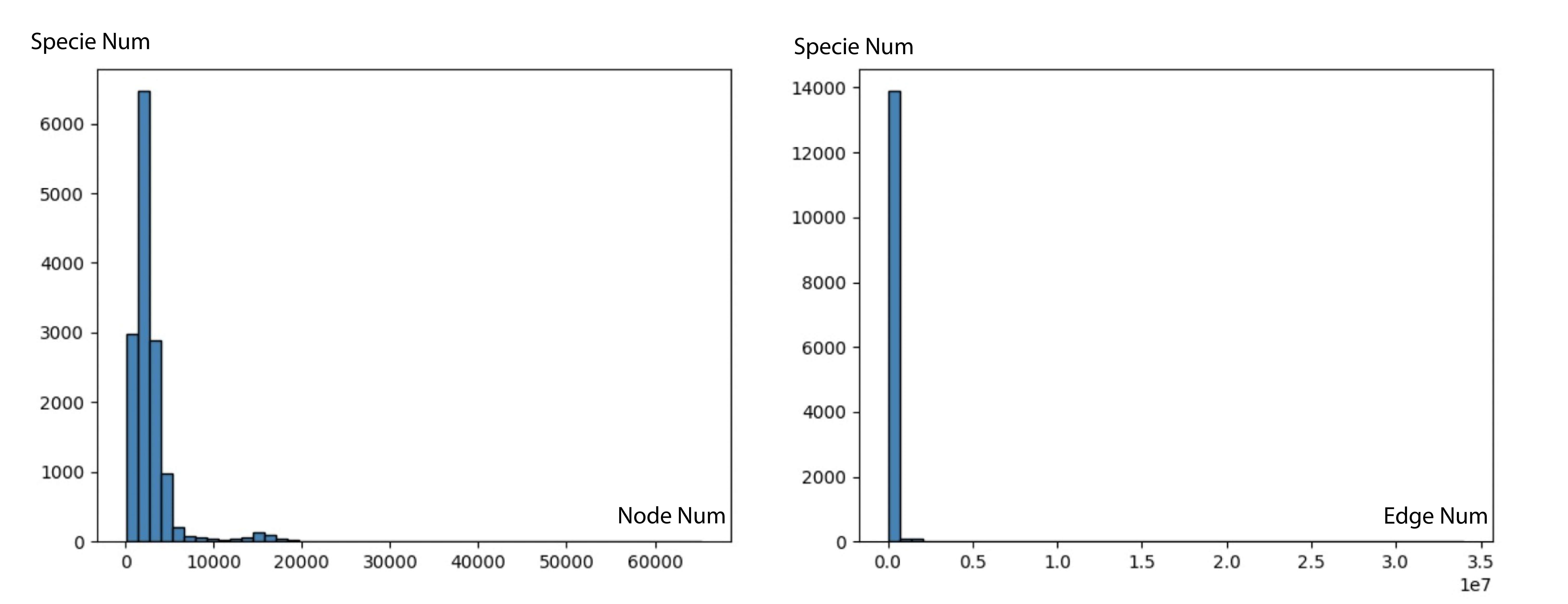}}
\caption{Visualization of the number of nodes and the number of edges in the original database.}
\label{origin-distribution}
\end{center}
\vskip -0.3in
\end{figure}

\begin{figure}[h]
%\vskip 0.2in
\begin{center}
\centerline{\includegraphics[width=\columnwidth]{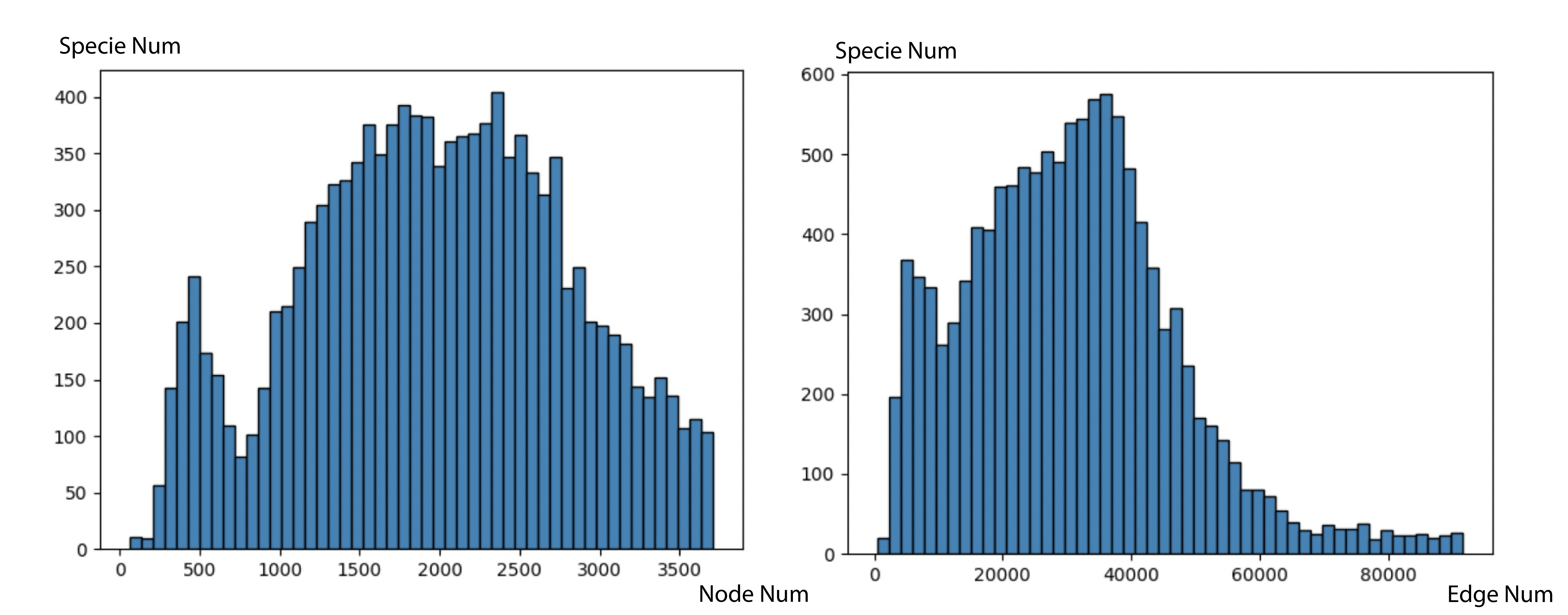}}
\caption{Visualization of the number of nodes and the number of edges in the pre-processed database.}
\label{process-distribution}
\end{center}
\vskip -0.3in
\end{figure}

We observe that the PPI network has a problem of uneven distribution, as illustrated in Figure~\ref{origin-distribution}, the largest network contains 60,000 proteins and $3.5\times10^7$ edges. Such data distributions can lead models to over-focus on proteins from a single species. We pre-process our dataset by choosing the species networks with comparable sizes. Figure~\ref{process-distribution} illustrates the data distribution after being pre-processed.

\vskip -0.1in
\begin{table}[h]
\caption{Statistics of the downstream datasets. EC., ER., TPC., and TC. respectively represent entry classification, entry regression, token pairwise classification, and token classification. }
\label{ppi-static}
\vskip -0.1in
\begin{center}
\begin{small}
\begin{sc}
\begin{tabular}{lrrc}
\toprule
Dataset & \#Protein & \#Entry & Task \\
\midrule
SHS27k & 1,690 & 7,624 & EC.\\
SHS148k & 5,189 &  44,488 & EC.\\
STRING-HomoSapiens & 15,335 & 593,397 & EC.\\
SAbDab & 493 & 1,479 & ER. \\
Proteinnet & 25,339 & 25,339 & TPC.\\
ICProtein & 2,212& 2,212& TPC.\\
CB513 & 3,078 & 3,078 &TC.\\
Fluorescence & 3,043 & 3,043 &ER.\\
Stability & 1,665 & 1,665 & ER.\\
\bottomrule
\end{tabular}
\end{sc}
\end{small}
\end{center}
\vskip -0.1in
\end{table}

\section{Datasets Statistics}

The statistical results of the dataset are shown in Table~\ref{ppi-static}. The number of proteins refers to the total number of occurrences in the training and test sets. An entry refers to a data point that contains input and output. Note that in some tasks (such as PPI prediction and antibody-antigen binding affinity prediction), an input contains more than one protein.
\section{ICProtein}
To construct a contact map dataset based on 3D interaction conformation, we firstly obtain all the data of protein complex structures from~\citeauthor{doi:10.1126/science.abm4805}\yrcite{doi:10.1126/science.abm4805}. 
Then we use obabel~\cite{obabel} to convert the CIF file to the PDB file~\cite{pdb2cm}. By calculating the distance of residues,  we obtain the contact map based on interaction conformation. 
Since each CIF file consists of two interacting proteins, we separate them and get 2,212 protein interaction contact maps.

\begin{figure}[h]
%\vskip 0.2in
\begin{center}
\centerline{\includegraphics[width=0.9\columnwidth]{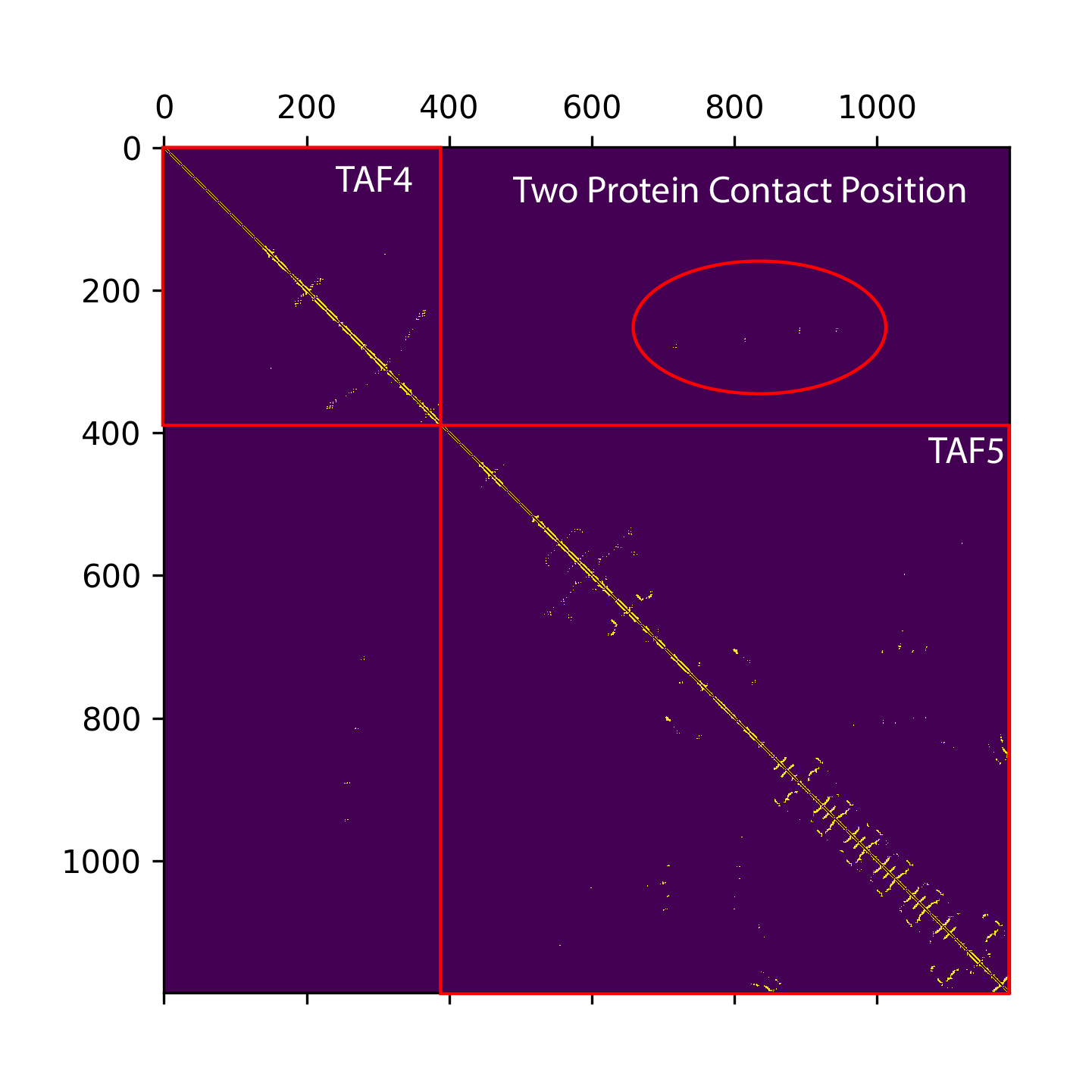}}
\caption{Visualization of TAF4 and TAF5 contact map.}
\label{TAF4-contact-map}
\end{center}
\vskip -0.3in
\end{figure}

In Figure~\ref{fig:ablation}, we illustrate the distance of corresponding amino acids with and without conformational information of TAF4 protein. Compared with the contact map (Figure~\ref{TAF4-contact-map}), we find that our method successfully captures the binding sites of TAF4 and TAF5. This finding indicates that the IC prompt can enhance protein representations by changing the embedding of PPI-related amino acids.

\begin{table*}
\caption{Hyper-parameter Search Space of Our Method}
\label{search-space}
\begin{center}
\begin{small}
\begin{sc}
\resizebox{.99\textwidth}{!}{
\begin{tabular}{lccccc}
\toprule
Dataset & batch\_size & max\_seq\_length & learning\_rate & weight\_decay & warmup\_updates  \\
\midrule
SHS27k & \{512, 1024, \textbf{2048}\} & 2,000 & \{1e-5, 5e-5, \textbf{1e-4}, 2e-4\} & \{0.0, 1e-4, \textbf{5e-4}\} & - \\
SHS148k & \{512, 1024, \textbf{2048}\} & 2,000 & \{1e-5, 5e-5, \textbf{1e-4}, 2e-4\} & \{0.0, 1e-4, \textbf{5e-4}\} & - \\
STRING-HomoSapiens & \{512, 1024, \textbf{2048}\} & 2,000 & \{1e-5, 5e-5, \textbf{1e-4}, 2e-4\} & \{0.0, 1e-4, \textbf{5e-4}\} & - \\
SAbDab & 1 & 1,024 &\{\textbf{1e-5}, 3e-5, 1e-4, 2e-4\} & - & \{\textbf{0}, 100, 10,000\}\\
CASP12 & 1 & 1,024 & \{1e-5, \textbf{3e-5}, 1e-4, 2e-4\} & - & \{\textbf{0}, 100, 10,000\} \\
ICProtein  & 1 & 1,024 & \{1e-5, \textbf{3e-5}, 1e-4, 2e-4\} & - & \{\textbf{0}, 100, 10,000\}\\
CB513 & 2 & 1,024 &\{1e-5, \textbf{3e-5}, 1e-4, 2e-4\} & - & \{\textbf{0}, 100, 10,000\}\\
Fluorescence & 8 & 512 & \{\textbf{1e-5}, 3e-5, 1e-4, 2e-4\}& - & \{0, 100, \textbf{10,000}\}\\
Stability & 2 & 1,024& \{\textbf{1e-5}, 3e-5, 1e-4, 2e-4\}&  - &\{0, 100, \textbf{10,000}\}\\
\bottomrule
\end{tabular}}
\end{sc}
\end{small}
\end{center}
\vskip -0.1in
\end{table*}

\section{Downstream Task Definition}
Here, we define a list of downstream tasks and their inputs and outputs.
\begin{itemize}
\item \textbf{Protein-Protein Interaction Prediction} is a sequence-level classification task. Its input contains two amino acid sequences $S_i, S_j$, and since STRING divides PPI into 7 categories, we need to predict the interactions and types between these two proteins $y_{ij} \in \{0, 1\}^7$.
\item \textbf{Antibody-Antigen Binding Affinity Prediction} is a sequence-level  regression task. Its input contains three amino acid sequences $S_i, S_j, S_k$, and we need to predict the value of binding affinity $y_{ijk} \in \mathbb{R}$.
\item \textbf{Contact Prediction} is a token-level classification task. Its input is an amino acid sequence $S=\{s^1, \cdots, s^n\}$, and we need to predict whether two residues are in contact $y \in \{0, 1\}^{n\times n}$.
\item \textbf{Secondary Structure Prediction} is a token-level classification task. Its input is an amino acid sequence $S=\{s^1, \cdots, s^n\}$, and we need to predict the type of each amino acid $y \in \{\mathtt{Helix}, \mathtt{Strand}, \mathtt{Other}\}^n$.
\item \textbf{Fluorescence} is a sequence-level regression task. Its input is an amino acid sequence $S$, and we need to predict the log-florescence intensity $y\in\mathbb{R}$.
\item \textbf{Stability} is a sequence level regression task. Its input is an amino acid sequence $S$, and we need to predict the most extreme circumstances in which $S$ maintains its activity $y\in\mathbb{R}$.
\end{itemize}

\section{Experimental Details}

%\subsection{Hyper-parameter Search Space of Our Method}
We present our hyper-parameter search space in Table~\ref{search-space}. The first three datasets are for the PPI prediction task. In this task, we freeze ConfProtein parameters and train the graph neural network predictor, so we only search the predictor parameters. The max sequence length $n$ denotes that the length of each input amino acid sequence will be padded by $\mathtt{[PAD]}$ to make totally $n$ tokens.

\end{document}